\documentclass{llncs}
\usepackage{amssymb}
\usepackage{amsmath}
\usepackage{multirow}
\usepackage{wrapfig}
\usepackage{hyperref}
\usepackage{graphicx}
\usepackage{makecell}
\begin{document}
\title{Complex Fully Convolutional Neural Networks for MR Image Reconstruction}
\titlerunning{Muneer Ahmad et al.}
% \titlerunning{Anonymous}  % abbreviated title (for running head)
\author{
Muneer Ahmad Dedmari\inst{1,2} \and Sailesh Conjeti\inst{1,2} \and Santiago Estrada \inst{1,2} \and Phillip Ehses\inst{1} \and Tony St\"{o}cker\inst{1}
\and Martin Reuter\inst{1,3}
}

% \author{Paper \# 224}
%

% \tocauthor{Anonymous}
%
% \institute{Submitted to MICCAI-MLMIR 2018}
\institute{
%Submitted to MICCAI-MLMIR 2018
%*************************************************************
German Center for Neurodegenrative Diseases (DZNE), Bonn, Germany \\
and
%*************************************************************
Computer Aided Medical Procedures, Technische Universit\"at M\"unchen, Germany\\
%\email{fernando.navarro@tum.de}
and
%*************************************************************
Harvard University and Massachusetts General Hospital, Boston, USA.
}
\maketitle              % typeset the title of the contribution
\begin{abstract}

Undersampling the \textit{k}-space data is widely adopted for acceleration of Magnetic Resonance Imaging (MRI). Current deep learning based approaches for supervised learning of MRI image reconstruction employ real-valued operations and representations by treating complex valued k-space/spatial-space as real values. In this paper, we propose complex dense fully convolutional  neural network ($\mathbb{C}$DFNet) for learning to de-alias the reconstruction artifacts within undersampled MRI images. We fashioned a densely-connected fully convolutional block tailored for complex-valued inputs by introducing dedicated layers such as complex convolution, batch normalization, non-linearities \textit{etc}. $\mathbb{C}$DFNet leverages the inherently complex-valued nature of input \textit{k}-space and learns richer representations. We demonstrate improved perceptual quality and recovery of anatomical structures through $\mathbb{C}$DFNet in contrast to its real-valued counterparts.

\end{abstract}
\section{Introduction}

Magnetic Resonance (MR) Imaging is widely adopted in many diagnostic applications due to its improved soft-tissue contrast, non-invasiveness and excellent spatial resolution. However, MRI is associated with long scan durations as the data is read out sequentially in \textit{k}-space and the speed at which the \textit{k}-space can be traversed is limited by the underlying imaging physics. This in turn limits the clinical use of MRI, causes inconvenience to patients, and renders this modality expensive and less accessible. One potential approach to accelerate MRI acquisition is to undersample \textit{k}-space \textit{i.e.} reduce the number of \textit{k}-space traversals made during acquisition. However, such an undersampling violates the Nyquist-Shannon Sampling theorem~\cite{nyquist} and generates aliasing artefacts upon reconstruction. A learning based reconstruction algorithm should effectively compensate for missing \textit{k}-space samples by leveraging \textit{a priori} knowledge of the anatomy at hand and the undersampling pattern. 

\begin{figure}
\vspace{-10pt}
\begin{center}
\includegraphics[width=\textwidth]{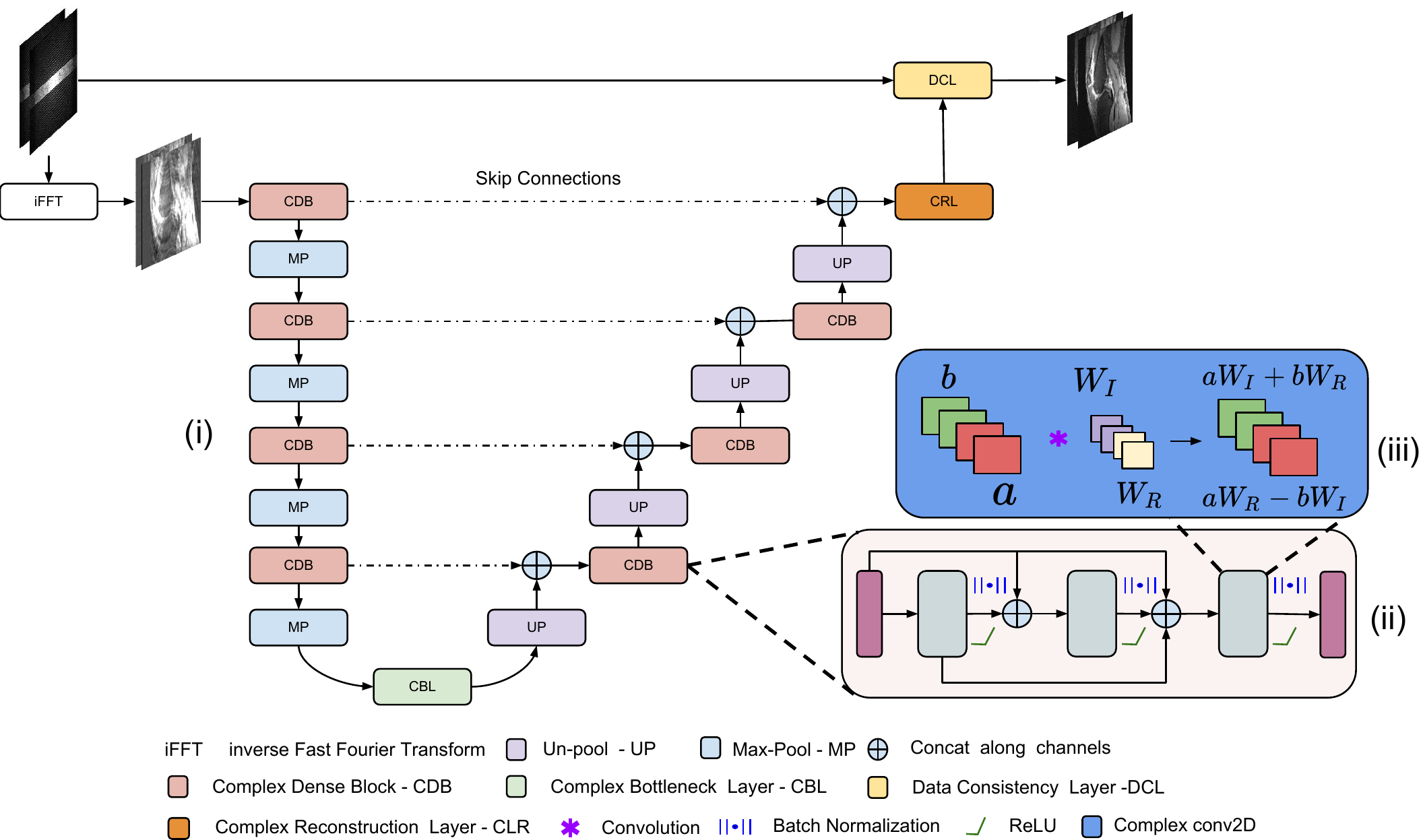}
\end{center}
\vspace{-20pt}
\caption{\small{(i). Complex Fully Convolutional Neural Network Architecture. (ii). Complex dense block, composed of 3 complex conv2D layers, followed by complex batch normalization and ReLU. (iii). Complex Conv2D layer, responsible for performing complex convolution operation, here ${a}$ and ${b}$ represents real and complex feature maps, and ${W_R}$ and ${W_I}$ represents real and imaginary parts of learnable weights.}}
\label{fig:architecture}
\vspace{-15pt}
\end{figure}

Deep learning is being increasingly adopted for MR reconstruction. Instead of using handcrafted features, Hammernik \textit{et al.}~\cite{variationalNetwork} demonstrated learning a set of regularizers under a variational framework, for reconstruction of accelerated MRI data. Kinam \textit{et al.}~\cite{perceptron} used the multilayer perceptron for accelerated parallel MRI. These works were further extended using techniques such as, deep residual learning~\cite{DeepRL}, domain adaptation~\cite{domainAdaptation}, data consistency layer~\cite{CascasedNet}, manifold approximation (AUTOMAP)~\cite{automap}, to name a few. However, all of the above mentioned reconstruction methods employ real-valued convolution operations in the spatial-domain by treating real (amplitude) and imaginary (phase) parts as two independent components. It should be noted that unlike multi-channel images (such as RGB images) where individual channels are acquired independently, MR data is inherently complex-valued in nature. Quadrature detection is employed to measure the changing circularly polarized magnetic field within the scanner which results in two simultaneously acquired data streams with a $\pi/2$ phase difference. Upon digitization, these signals constitute the real and imaginary parts of each complex data point in the \textit{k}-space. The magnitude derived from this complex valued data mainly carries information about proton density as well as relaxation properties of the tissue. The phase can be used to obtain the information, for example, about magnetic susceptibility, flow, or temperature. To faithfully recover the complete \textit{k}-space, it is important to learn the co-relationship between these data-streams.

In this paper, for the first time, we explore end-to-end learning with complex-valued data targeted at MR reconstruction. Towards this, we propose the Complex Dense Fully Convolutional Network ($\mathbb{C}$DFNet) by introducing densely connected fully convolutional blocks made with layers supporting deep learning operations on complex valued data. Complex-valued arithmetic operators for deep learning were proposed by Trabelsi \textit{et al.}~\cite{complexNet} where complex counterparts of convolution, batch-normalization, network initialization \textit{etc.} were explored. We also propose a composite loss function that simultaneously minimizes reconstruction error while improving structural similarity.

\section{Methodology}

\subsection{Problem Formulation}
Let the fully-sampled complex-valued MR image be represented as ${\mathbf{x}_{f}\in \mathbb{C}^N }$ consisting of ${\sqrt{N}\times\sqrt{N}}$ pixels arranged in a column fashion with each pixel composed of a complex vector with real and imaginary components. This image is reconstructed from fully-sampled measurements in \textit{k}-space, say ${\mathbf{y}_{f}\in \mathbb{C}^N }$, such that: $\mathbf{y}_{f} = \mathbf{F}_{f}\mathbf{x}_{f}$, where $\mathbf{F}_{f} \in \mathbb{C}^{N \times N}$ is the fully sampled encoding matrix. During under-sampling, we acquire measurements in \textit{k}-space, say $\mathbf{y}_{u} \in \mathbb{C}^{M}$ where $M \ll N$. Let the image reconstructed from zero-filling $\mathbf{y}_{u}$ be represented as $\mathbf{x}_{u}$, such that $\mathbf{x}_{u} = \mathbf{F}_{u}^{-1}\mathbf{y}_{u}$. 
Reconstructing $\mathbf{x}_{f}$ directly from $\mathbf{y}_{u}$ is ill-posed and direct inversion is not possible due to the under-determined nature of the system of equations. In our approach, we enforce $\mathbf{x}_{f}$ to be approximated using a complex fully convolutional neural network (represented as $f_{\mathbb{C}}$). As $\mathbf{x}_{u}$ is highly-aliased due to sub-Nyquist sampling, $f_{\mathbb{C}}$ aims at recovering image $\mathbf{x_r}$ that is as close as possible to an ideal fully sampled image $\mathbf{x}_{f}$. 

\subsection{Network Architecture}

%Input to the network is complex valued undersampled image, where real and imaginary parts are stored in two channels i.e. $\mathbb{R}^{\sqrt{N}\times\sqrt{N}\times2}$. \mathbin{/} 

%\noindent
\textbf{Complex Dense Block}: The densely connected block proposed in~\cite{densenet}, introduces feed-forward connections from each layer to every other layer (illustrated in Fig.~\ref{fig:architecture}(ii)). Such an architecture choice was demonstrated to encourage feature reusability and strengthen information propagation through the network. We suitably adapt this block for complex valued data by proposing counterparts of classic deep learning layers such as convolution, batch normalization, non-linearity (ReLU), up-sampling \textit{etc.} For sake of brevity, we delve only into the complex convolution (denoted as $*_{\mathbb{C}}$) in detail. Let $\mathbf{h} = \mathbf{a} + i\mathbf{b}$ be the complex-valued input to convolution layer with weights $\mathbf{W} = \mathbf{W_R} + i \mathbf{W_I}$, the complex convolution between $\mathbf{h}$ and $\mathbf{W}$ is simulated using real-valued arithmetic as: $\mathbf{W} *_{\mathbb{C}} \mathbf{h} = \left ( \mathbf{a} * \mathbf{W_R} - \mathbf{b} * \mathbf{W_I} \right ) + i\left ( \mathbf{a} * \mathbf{W_I} + \mathbf{b} * \mathbf{W_R}  \right )$, as shown in Fig.~\ref{fig:architecture}(iii). The complex output feature maps are fed into the complex batch normalization layer, which normalizes the data to have equal variance along the real and imaginary components, thereby ensuring a co-relationship between them. The complex variant of non-linearity ReLU and max-pooling are applied on the real and imaginary channels separately.

\noindent
\textbf{Complex Dense Fully Convolutional Network ($\mathbb{C}$DFNet)}: The $\mathbb{C}$DFNet $f_{\mathbb{C}}$ is based on the DenseNet~\cite{densenet} architecture, comprising of a sequence of four densely-connected complex encoder blocks with corresponding densely-connected complex decoder blocks separated by a bottleneck layer (illustrated in Fig.~\ref{fig:architecture}(i). The output of the last decoder block is given to a reconstruction layer (with complex convolution operators) for reconstructing the image. The encoders and decoders are stacked and trained in a progressive way \textit{i.e.} output from one block is used as input to other block. Skip connections are included in the architecture between encoder and corresponding decoder blocks to fuse high-level representations (decoder) with low-level features (encoder) for preserving contextual information. Furthermore, skip connections prevent the vanishing gradient problem, by directly propagating gradients from decoder to respective encoder block.
%Skip connections are included in the architecture between encoder and corresponding decoder blocks of same spatial resolution. Skip connections from encoder block to decoder block fuses high-level representations(decoder) with low-level features(encoder) for preserving contextual representation and helps to prevent the vanishing gradient problem, by directly propagating gradients from the decoder block to the respective encoder block~\cite{Unet}.% 
The network $f_{\mathbb{C}}$ takes complex-valued aliased image $\mathbf{x}_{u}$ (generated by zero-filling under-sampled \textit{k}-space data $\mathbf{y}_{u}$)  as input to an intermediate reconstructed image $\widetilde{\mathbf{x}}_{r}$ which is fed further into the data consistency layer for imputing missing \textit{k}-space values.

\noindent
\textbf{Data Consistency Layer (DCL)}: We recover a full reconstructed \textit{k}-space spectrum $\widetilde{\mathbf{y}}_{r}$ via a Fourier transform on the reconstructed image $\widetilde{\mathbf{x}}_{r}$. To retain all the \textit{a priori} available \textit{k}-space values $\mathbf{y}_{u}$ (collected at spatial locations denoted \textit{via} mask $\Omega$) and impute only the missing values at locations ($\not\in \Omega$), the data consistency layer performs the following operation: 
\begin{equation}
\mathbf{y}_{r}\left ( z \right ) = \left\{\begin{matrix}
\mathbf{y}_{u}\left ( z \right ) & z \in \Omega \\ 
\widetilde{\mathbf{y}}_{r}\left ( z \right ) & z \not\in \Omega
\end{matrix}\right.
\end{equation}
\noindent
After the DCL layer, the final de-aliased image $\mathbf{x}_{r}$ is recovered through inverse Fourier transform of $\mathbf{y}_{r}$. It must be noted that the inclusion of the DCL layer within $f_{\mathbb{C}}$ ensures improved efficacy of the network by focusing exclusively on missing \textit{k}-space values and enforces consistency with \textit{a priori} acquired data $\mathbf{y}_{u}$. Further, the DCL layer does not have any learnable parameters and does not increase the complexity of the network.

%While reconstruction, the network not only recovers the missing data, but also modifies the actual data from the ${K_U}$. The data consistency layer i.e. Fig.~\ref{fig:dcl},  is replacing the modified values by ${M}$ non-zero entries of ${K_U}$ i.e. k-space correction[reference to Deep learning for undersampled MRI reconstruction]. In order to perform k-space correction, data is first transformed from 2 channeled image-space to complex k-space by Fourier transform. After correction of k-space, the data is transformed back to 2 channeled image-space by inverse Fourier transform and storing real and imaginary parts in 2 channels. By including  data-consistency layer in training pipeline, it improves the efficiency of network by focusing only on missing entries. Data consistency layer do not have any learnable parameter, hence not increasing the complexity of the network architecture.

%\[
%\begin{bmatrix}
%   x_{11}       & x_{12} & x_{13} & \dots & x_{1n} \\
%   x_{21}       & x_{22} & x_{23} & \dots & x_{2n} \\
%    \hdotsfor{5} \\
%    x_{d1}       & x_{d2} & x_{d3} & \dots & x_{dn}
%\end{bmatrix}
%=
%\begin{bmatrix}
%    x_{11} & x_{12} & x_{13} & \dots  & x_{1n} \\
%    x_{21} & x_{22} & x_{23} & \dots  & x_{2n} \\
%    \vdots & \vdots & \vdots & \ddots & \vdots \\
%    x_{d1} & x_{d2} & x_{d3} & \dots  & x_{dn}
%\end{bmatrix}
%\]

\subsection{Model Learning and Optimization}
The network $f_{\mathbb{C}}$ is optimized to recover missing \textit{k}-space data while simultaneously preserving fine-grained anatomical details. We adopt a supervised learning approach wherein a training dataset $\mathcal{D}$ of input-target (under-sampled and fully-sampled) pairs ($\mathbf{x}_{u},\mathbf{x}_{f}$) to train $f_{\mathbb{C}}$. We use a composite loss function comprising of two contributing terms, firstly a mean-squared error term (${\mathcal{L}_{L_{2}}}$) and secondly Structural Similarity Index Measure (SSIM) ($\mathcal{L_{\text{SSIM}}}$) as discussed below: 

\noindent
\textbf{$\mathcal{L}_{L_{2}}$ Loss}: This loss is used to minimize the difference between the reconstructed image $\mathbf{x}_{r}$ and target fully sampled image $\mathbf{x}_{f}$. 
\begin{equation}
\mathcal{L}_{L_{2}} = \sum_{\left ( \mathbf{x}_{u},\mathbf{x}_{f} \right ) \in \mathcal{D}} \left \| \mathbf{x}_{f} - \mathbf{x}_{r} \right \|^{2}_{2} = \sum_{\left ( \mathbf{x}_{u},\mathbf{x}_{f} \right ) \in \mathcal{D}} \left \| \mathbf{x}_{f} - f_{\mathbb{C}}\left ( \mathbf{x}_{u} | \theta \right ) \right \|^{2}_{2} 
\end{equation}
\noindent
The $\mathcal{L}_{2}$ loss penalizes large errors, but fails to capture finer details which the human visual system is sensitive to such as contrast, luminance and structure. To offset the above shortcoming of $\mathcal{L}_{2}$ loss, we use SSIM~\cite{ssim}, which is perceptually closer to the human visual system, as an additional loss $\mathcal{L_{\text{SSIM}}}$, defined as: 
\begin{equation}
\mathcal{L_{\text{SSIM}}} = \sum_{\left ( \mathbf{x}_{u},\mathbf{x}_{f} \right ) \in \mathcal{D}}\left ( 1 - \mathcal{S}\left ( \mathbf{x}_{r},\mathbf{x}_{f} \right ) \right )
\end{equation}
\noindent
where $\mathcal{S}\left ( \mathbf{x}_{r},\mathbf{x}_{f} \right ) $ is the SSIM calculated between $\mathbf{x}_{r}$ and $\mathbf{x}_{f}$. The composite loss function $\mathcal{L}$ for optimizing $f_{\mathbb{C}}$ is defined as: $\mathcal{L}\left ( \mathbf{x} , f_{\mathbb{C}}\left ( \mathbf{x}_{u} | \theta \right ) \right ) = \mathcal{L}_{L_{2}} + \lambda \mathcal{L_{\text{SSIM}}}$, where ${\lambda}$ is a scaling constant.

\begin{figure} [t!]
\vspace{-15pt}
\begin{center}
\includegraphics[width=\textwidth]{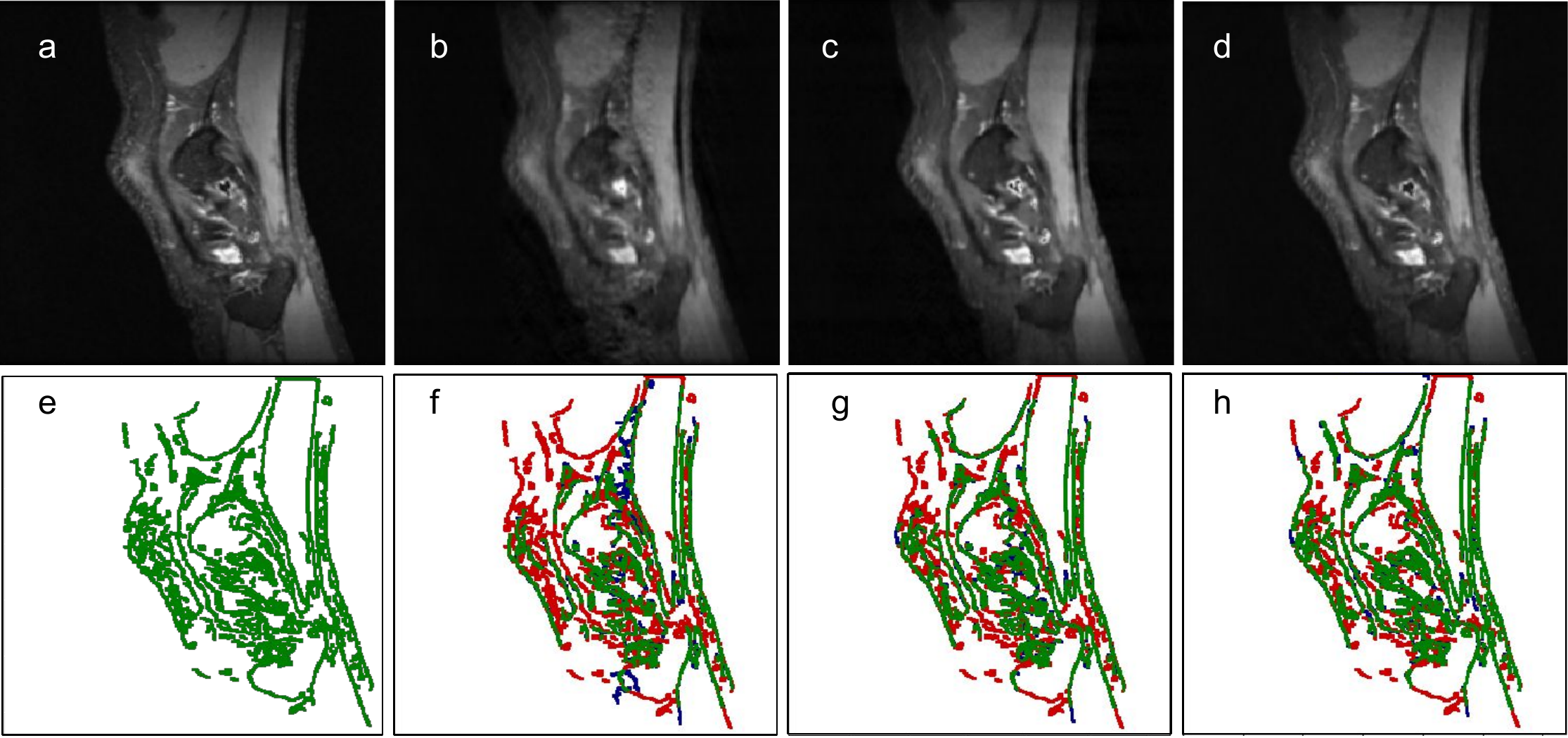}
\end{center}
\vspace{-17pt}
\caption{\small{Edge-map results comparison at undersampling factor of x4. (a), (e) ground-truth and its edge-map, (b),(f) undersampled, and its edge-map (c), (g) DLMRI reconstruction and its edge-map  (d), (h) proposed reconstruction and its edge-map. Here, green represents edges present in ground-truth, red represents edges that are missing in reconstructed image, as compared to ground-truth and blue represents edges that are not present in ground-truth but only in reconstructed images. }}
\label{fig:results_fom}
\vspace{-15pt}
\end{figure}

\section{Results and Discussion}
\subsection{Experimental Settings and Evaluation}
\label{ss:exptsettings}
\textbf{Dataset}
Our experiments were evaluated on the publicly available 20 fully-sampled knee \textit{k}-space dataset from \href{http://mridata.org/}{mridata.org}~\cite{mridata}. The data was split randomly into 16 patients for training and rest for testing. The coils were fused using sum of squares into a single complete \textit{k}-space dataset and training data for proof-of-concept was generated using Cartesian under-sampling proposed in~\cite{CascasedNet}, wherein eight lowest spatial frequencies were preserved and a zero-mean Gaussian distribution was used to determine the sampling probability along the phase encoding direction (the frequency-encoding direction was fully-sampled). 

\noindent
\textbf{Baselines and Comparative Methods}: To ablatively test the introduction of complex convolution, we compare with the na\"{i}ve variant of densely connected networks treating the complex-valued input as two independent channels (termed BL1). We further compare the contribution of the data-consistency layer by defining a variant sans DCL (termed BL2). Finally, to evaluate the contribution of training with $\mathcal{L_{\text{SSIM}}}$, we set the corresponding factor $\lambda$ to 0 and contrast with the proposed method (termed BL3). Further, we compare against a state-of-the art dictionary learning based MR reconstruction method proposed in~\cite{dlmri} (termed as DLMRI). It must be noted that BL1 is akin to deep learning based reconstruction method proposed in~\cite{unet_recon}, differing only in the usage of densely-connected blocks. In all the aforementioned network configurations, we used complex convolution operators (except BL1) with a depth of 32, and kernel size of $3 \times 3$, BL1 was designed with depth of 46 for a fair comparison in terms of the number of learnable parameters. The networks were trained until convergence using RMSProp as an optimizer with a learning rate of ${5e^{-5}}$ with decay of 0.9 and batch-size of 5 for 50 epochs. %The networks were implemented with TensorFlow framework and were tra

The networks were evaluated at two acceleration factors of 4$\times$ and 6$\times$ along the phase-encoding directions. During training of the deep networks, the under-sampling masks were generated on-the-fly to induce the tolerance towards a range of potential aliasing artefacts. We further used image-level rigid and elastic transformations to augment the training data. As demonstrated in~\cite{CascasedNet}, fidelity of image reconstruction is evaluated by measuring the similarity between a reconstructed image to the fully-sampled ground truth image using metrics such as SSIM, mean squared error (MSE) \textit{etc.} However, these metrics do not explicitly focus on finer details of the reconstruction and towards this we employ Pratt's figure of merit (Pratt's FOM)~\cite{PFOM} as an additional metric. Pratt's FOM exclusively focuses on the edges and corner points present in the reconstructed image that are concurrent with structures present in the ground truth image while simultaneously penalizing both missing and artificially hallucinated edges.

\subsection{Results}
\begin{wraptable}[13]{r}{0.55\textwidth}
%\begin{table}[]
\vspace{-35pt}
\centering
\caption{Pratt's Figure of Merit of comparative analysis against baselines}
\label{baselines}
\vspace{2pt}
\resizebox{0.53\textwidth}{!}{
\begin{tabular}{|c|c|c|c|c|c|}
\hline
\multicolumn{2}{|c|}{\textbf{Acceleration}} & \multicolumn{4}{c|}{\textbf{Pratt's FOM}}                     \\ \hline
\textbf{Train}        & \textbf{Test}       & \textbf{BL1} & \textbf{BL2} & \textbf{BL3} & \textbf{Proposed}     \\ \hline
4$\times$                    & 4$\times$                  & 0.81657      & 0.77522      & 0.82480      & \textbf{0.84364} \\ \hline
6$\times$                    & 4$\times$                  & 0.83961      & 0.7743       & 0.82409      & \textbf{0.84218} \\ \hline
4$\times$                    & 6$\times$                  & 0.71775      & 0.70099      & 0.75155      & \textbf{0.77449} \\ \hline
6$\times$                    & 6$\times$                  & 0.76009      & 0.7199       & 0.75661      & \textbf{0.77514} \\ \hline
\multicolumn{6}{|l|}{\textbf{BL1}: DenseNet with $\lambda = 2$ with DCL} \\
\multicolumn{6}{|l|}{\textbf{BL2}: $\mathbb{C}$DFNet with $\lambda = 2$ without DCL} \\
\multicolumn{6}{|l|}{\textbf{BL3}: $\mathbb{C}$DFNet with $\lambda = 0$ with DCL} \\ \hline
\end{tabular}
}
%\vspace{-10pt}
% \textbf{BL${_1}$}: DenseNet with ${\lambda}$=2 and using DCL, \textbf{BL${_2}$}: Complex DenseNet with ${\lambda}$=2 and not using DCL, \textbf{BL${_3}$}: Complex DensNet with ${\lambda}$=0 and using DCL, and \textbf{Proposed}: C-FCNN
%\end{table}
\end{wraptable}
The networks trained for 4$\times$ and 6$\times$ acceleration factors were tested across and within these factors resulting in four train-test combinations. All the methods were evaluated for each of these combinations to quantify their generalizability to unseen aliasing effects.

\begin{table}[]
\vspace{-10pt}
\centering
\caption{Quantitative Comparison from Cartesian trajectory with undersampling factor of 4$\times$ and 6$\times$}
\vspace{-7pt}
\label{dlmri_results}
\resizebox{\textwidth}{!}{
\begin{tabular}{|c|c|c|c|c|c|c|c|c|c|c|}

\hline
\multicolumn{2}{|c|}{\textbf{Acceleration}} & \multicolumn{3}{c|}{\textbf{SSIM}}                                    & \multicolumn{3}{c|}{\textbf{MSE (x${10^{-4}}$)}} & \multicolumn{3}{c|}{\textbf{Pratt's FOM}}                               \\ \hline
\textbf{Train}        & \textbf{Test}       & $\mathbf{x}_{u}$           & \textbf{DLMRI}          & \textbf{Proposed} & $\mathbf{x}_{u}$          & \textbf{DLMRI}          & \textbf{Proposed} & $\mathbf{x}_{u}$           & \textbf{DLMRI}           & \textbf{Proposed} \\ \hline
4$\times$                    & 4$\times$                  & \multirow{2}{*}{0.8886} & \multirow{2}{*}{0.9173} & \textbf{0.9269}   & \multirow{2}{*}{11.89}  & \multirow{2}{*}{7.01}   & \textbf{5.54}     & \multirow{2}{*}{0.63795} & \multirow{2}{*}{0.73876} & \textbf{0.84364}  \\ \cline{1-2} \cline{5-5} \cline{8-8} \cline{11-11} 
6$\times$                    & 4$\times$                  &                         &                         & 0.9266            &                         &                         & 5.57              &                          &                          & 0.84218           \\ \hline
4$\times$                    & 6$\times$                  & \multirow{2}{*}{0.8552} & \multirow{2}{*}{0.8920} & 0.9062            & \multirow{2}{*}{17.55}  & \multirow{2}{*}{10.70}  & 7.76              & \multirow{2}{*}{0.51309} & \multirow{2}{*}{0.64529} & 0.77449           \\ \cline{1-2} \cline{5-5} \cline{8-8} \cline{11-11} 
6$\times$                    & 6$\times$                  &                         &                         & \textbf{0.9072}   &                         &                         & \textbf{7.54}     &                          &                          & \textbf{0.77514}  \\ \hline
\end{tabular}}
\vspace{-15pt}
\end{table}

\begin{figure}[t!]
\vspace{-15pt}
\begin{center}
\includegraphics[width=\textwidth]{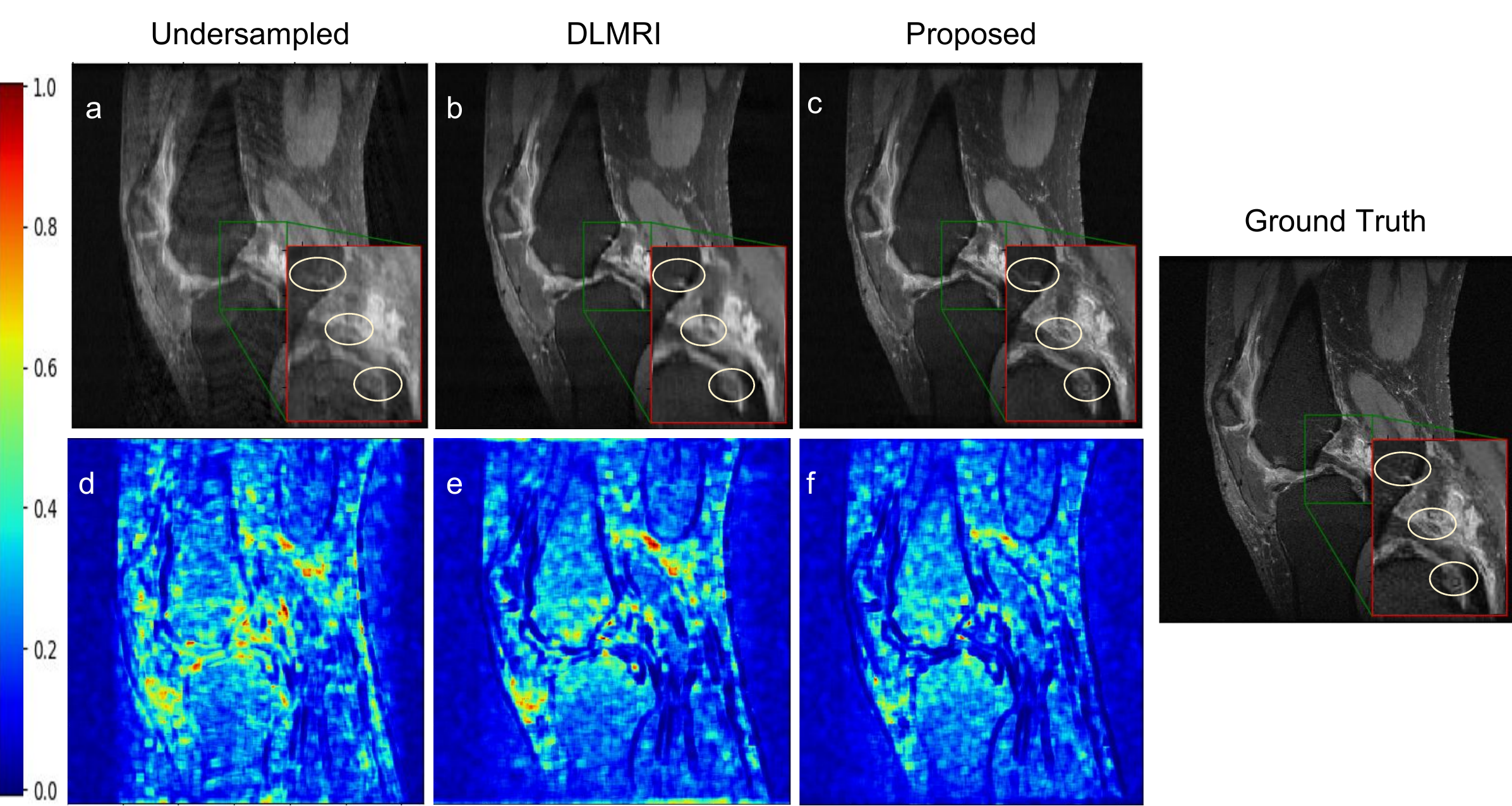}
\end{center}
\vspace{-15pt}
\caption{\small{Reconstruction results using 4$\times$ acceleration factor. (a), (d) Undersampled image and its error map, (b), (e) DLMRI reconstruction and its error map, (c), (f) Proposed reconstruction and its error-map, and ground truth.}}
\label{fig:results_dlmri}
\vspace{-17.5pt}
\end{figure}

\noindent
\textbf{Qualitative Analysis}: Fig.~\ref{fig:results_fom} illustrates the contrastive results on recovery of fine-grained details using the edge-map extracted from an under-sampled image (Fig.~\ref{fig:results_fom}(b,f)), DLMRI (Fig.~\ref{fig:results_fom}(c,g)) and proposed method (Fig.~\ref{fig:results_fom}(d,h)). We observe that the proposed network demonstrates maximal consistency in finer details with respect to the ground-truth. Fig.~\ref{fig:results_dlmri} highlights the differences with respect to the ground truth through a difference map and particularly focus on reconstruction of fine details in the region between the tibia and femur and the synovial membrane. 

\noindent
\textbf{Ablative Testing}: To ablatively evaluate the contributions of this work, the proposed method was contrasted against baselines (discussed in Sec.~\ref{ss:exptsettings}) and observations are tabulated in Table~\ref{baselines}. For sake of brevity, we only present the Pratt's FOM metric in this table. Contrasting the proposed method against BL1 in Table~\ref{baselines}, we observe a consistent improvement in the reconstruction error due to the introduction of complex dense blocks in place of vanilla dense blocks. This is particularly evident for the case of aggressive under-sampling (6$\times$) where the proposed method outperformed BL1 with a significant margin of 5.7\%. Comparing BL2 with the proposed method, the inclusion of the data consistency layer proved to be of high significance as evidenced across all validation combinations with an average improvement of over 6\%. The use of SSIM as an additional loss function during optimization (comparing BL3 with proposed method) also consistently improves Pratt's FOM across all the test cases. 

\noindent
\textbf{Comparative Methods}: In Table~\ref{dlmri_results}, we compare the proposed method against the under-sampled input image ($\mathbf{x}_{u}$) and state-of-art compressed sensing approach, DLMRI, in terms of the evaluation metrics SSIM, MSE and Pratt's FOM. We observe consistent improvement across all metrics in comparison to DLMRI, with the proposed method being able to recover finer details significantly (over 11\% improvement in Pratt's FOM). In scenarios of testing on aggressive acceleration (6$\times$), which corresponds to the limit of sparsity based methods, we observe that $\mathbb{C}$DFNet recovers anatomical details better as it is learnt in an end-to-end fashion allowing for efficient learning of anatomical priors from the training data.

\section{Conclusion and Future Work}
We have presented a deep learning based MR imaging reconstruction method, wherein real-valued neural network operations are replaced by complex convolutional operations. In this work, we demonstrated that the proposed network architecture outperformed the standard state-of art and the real-valued counter part methods by significant margins in terms of recovering fine structures and high frequency textures. The experiments also show that the proposed method is robust towards the undersampling ratio, which eliminates the need for training multiple large networks for each acquisition settings. Finally, Pratt's figure of merit was adapted for performing evaluation by considering the overall perceptual quality of reconstructed image. As {\it k}-space is inherently complex-valued, we believe that this method can be adapted to learn both, domain transformation as well as reconstruction. Moreover, non-Cartesian trajectories can be investigated, as they possess different aliasing properties, a further validation is appropriate to determine the flexibility of our method towards this end.

% \section{Acknowledgment}
% We are grateful to ....... for their financial support -- JUST TO CHECK NUMBER OF PAGES :)

\bibliographystyle{splncs04}
\bibliography{references}

\begin{thebibliography}{10}
\providecommand{\url}[1]{\texttt{#1}}
\providecommand{\urlprefix}{URL }
\providecommand{\doi}[1]{https://doi.org/#1}

\bibitem{PFOM}
Hagara, M., Hlavatovic, A.: Video segmentation based on pratt's figure of
  merit. In: 2009 19th International Conference Radioelektronika. pp. 91--94
  (April 2009). \doi{10.1109/RADIOELEK.2009.5158758}

\bibitem{densenet}
Huang, G., Liu, Z., Weinberger, K.Q.: Densely connected convolutional networks.
  CoRR  \textbf{abs/1608.06993} (2016)

\bibitem{unet_recon}
Hyun, C.M., Kim, H.P., Lee, S.M., Lee, S., Seo, J.K.: Deep learning for
  undersampled mri reconstruction. Physics in Medicine and Biology  (2018)

\bibitem{variationalNetwork}
Kerstin, H., Teresa, K., Erich, K., P., R.M., K., S.D., Thomas, P., Florian,
  K.: Learning a variational network for reconstruction of accelerated mri
  data. Magnetic Resonance in Medicine  \textbf{79}(6),  3055--3071

\bibitem{perceptron}
Kinam, K., Dongchan, K., HyunWook, P.: A parallel mr imaging method using
  multilayer perceptron. Medical Physics  \textbf{44}(12),  6209--6224

\bibitem{DeepRL}
Lee, D., Yoo, J.J., Tak, S., Ye, J.C.: Deep residual learning for accelerated
  mri using magnitude and phase networks. CoRR  \textbf{abs/1804.00432} (2018)

\bibitem{nyquist}
Nyquist, H.: Certain topics in telegraph transmission theory. Transactions of
  the American Institute of Electrical Engineers  \textbf{47}(2),  617--644
  (1928)

\bibitem{dlmri}
Ravishankar, S., Bresler, Y.: Mr image reconstruction from highly undersampled
  k-space data by dictionary learning. IEEE Transactions on Medical Imaging
  \textbf{30}(5),  1028--1041 (2011)

\bibitem{mridata}
Sawyer, A.M., Lustig, M., Alley, M., Uecker, P., Virtue, P., Lai, P.,
  Vasanawala, S., Healthcare, G.: Creation of fully sampled mr data repository
  for compressed sensing of the knee (2013)

\bibitem{CascasedNet}
Schlemper, J., Caballero, J., Hajnal, J.V., Price, A.N., Rueckert, D.: A deep
  cascade of convolutional neural networks for {MR} image reconstruction. CoRR
  \textbf{abs/1703.00555} (2017)

\bibitem{complexNet}
Trabelsi, C., Bilaniuk, O., Serdyuk, D., Subramanian, S., Santos, J.F., Mehri,
  S., Rostamzadeh, N., Bengio, Y., Pal, C.J.: Deep complex networks. CoRR
  \textbf{abs/1705.09792} (2017)

\bibitem{ssim}
Wang, Z., Bovik, A.C., Sheikh, H.R., Simoncelli, E.P.: Image quality
  assessment: from error visibility to structural similarity. IEEE Transactions
  on Image Processing  \textbf{13}(4),  600--612 (2004)

\bibitem{domainAdaptation}
Yoseob, H., Jaejun, Y., Hee, K.H., Jung, S.H., Kyunghyun, S., Chul, Y.J.: Deep
  learning with domain adaptation for accelerated projection‐reconstruction
  mr. Magnetic Resonance in Medicine  \textbf{80}(3),  1189--1205

\bibitem{automap}
Zhu, B., Liu, J.Z., Cauley, S.F., Rosen, B.R., Rosen, M.S.: Image
  reconstruction by domain-transform manifold learning. Nature  \textbf{555},
  487 EP -- (Mar 2018)

\end{thebibliography}

\end{document}